\documentclass[lettersize,journal]{IEEEtran}
\usepackage{amsmath,amsfonts}
\usepackage{algorithmic}
\usepackage{algorithm}
\usepackage{array}
\usepackage[caption=false,font=normalsize,labelfont=sf,textfont=sf]{subfig}
\usepackage{textcomp}
\usepackage{stfloats}
\usepackage{url}
\usepackage{verbatim}
\usepackage{graphicx}
\usepackage{cite}
\usepackage{booktabs}
\usepackage{multirow}
\begin{document}

\title{AD-H: Language-guided Autonomous Driving with Hierarchical Agents}

\author{
    Zaibin~Zhang$^*$, 
    Talas~Fu$^*$, 
    Shiyu~Tang$^*$, 
    Yuanhang~Zhang$^*$, \\
    Yifan~Wang,~\IEEEmembership{Member,~IEEE} ,
    Lijun~Wang$^\dagger$,~\IEEEmembership{Member,~IEEE}, 
    Huchuan~Lu,~\IEEEmembership{Fellow,~IEEE}\\
    \thanks{
        Zaibin~Zhang, Yuanhang~Zhang, Yifan Wang, Lijun Wang, Huchuan Lu is with Dalian University of Technology, Dalian, China (e-mail: zhangzaibin@mail.dlut.edu.cn; ljwang@dlut.edu.cn)
        }
    \thanks{
        $^*$Zaibin~Zhang, Talas~Fu, Shiyu~Tang and Yuanhang~Zhang contributed equally to this work. $^\dagger$Lijun Wang is the corresponding author.
        }
}

\maketitle

\begin{abstract}
Language-guided autonomous driving requires bridging a large abstraction gap between high-level natural-language instructions and low-level vehicle control. End-to-end approaches that use a single multimodal large language model (MLLM) to map language directly to actions struggle with this mismatch, often failing to exploit the model’s reasoning capabilities and exhibiting limited generalization beyond the distributions of driving datasets used for fine-tuning.
To address this issue, we propose AD-H, a hierarchical multi-agent framework that explicitly separates high-level decision-making from low-level vehicle execution. At the upper level, an MLLM-based planner interprets natural-language commands and environmental context to generate coherent mid-level driving instructions. At the lower level, a lightweight controller converts these mid-level instructions into precise, continuous control actions. This decomposition aligns with the functional strengths of each component: the planner focuses on semantic reasoning and task decomposition, while the controller ensures stable and accurate actuation.
To support large-scale training under this hierarchy, we design a rule-based pipeline that reconstructs mid-level commands from driving signals, producing 1.15 million hierarchical annotation pairs. Extensive experiments show that AD-H outperforms state-of-the-art models despite using fewer parameters (3B+350M vs. 7B), and achieves superior long-horizon generalization and instruction-following performance. We will make our data and code publicly accessible at \url{https://github.com/zhangzaibin/AD-H}
\end{abstract}


\section{Introduction}
\begin{figure*}[t]
\centering
\includegraphics[width=\textwidth]{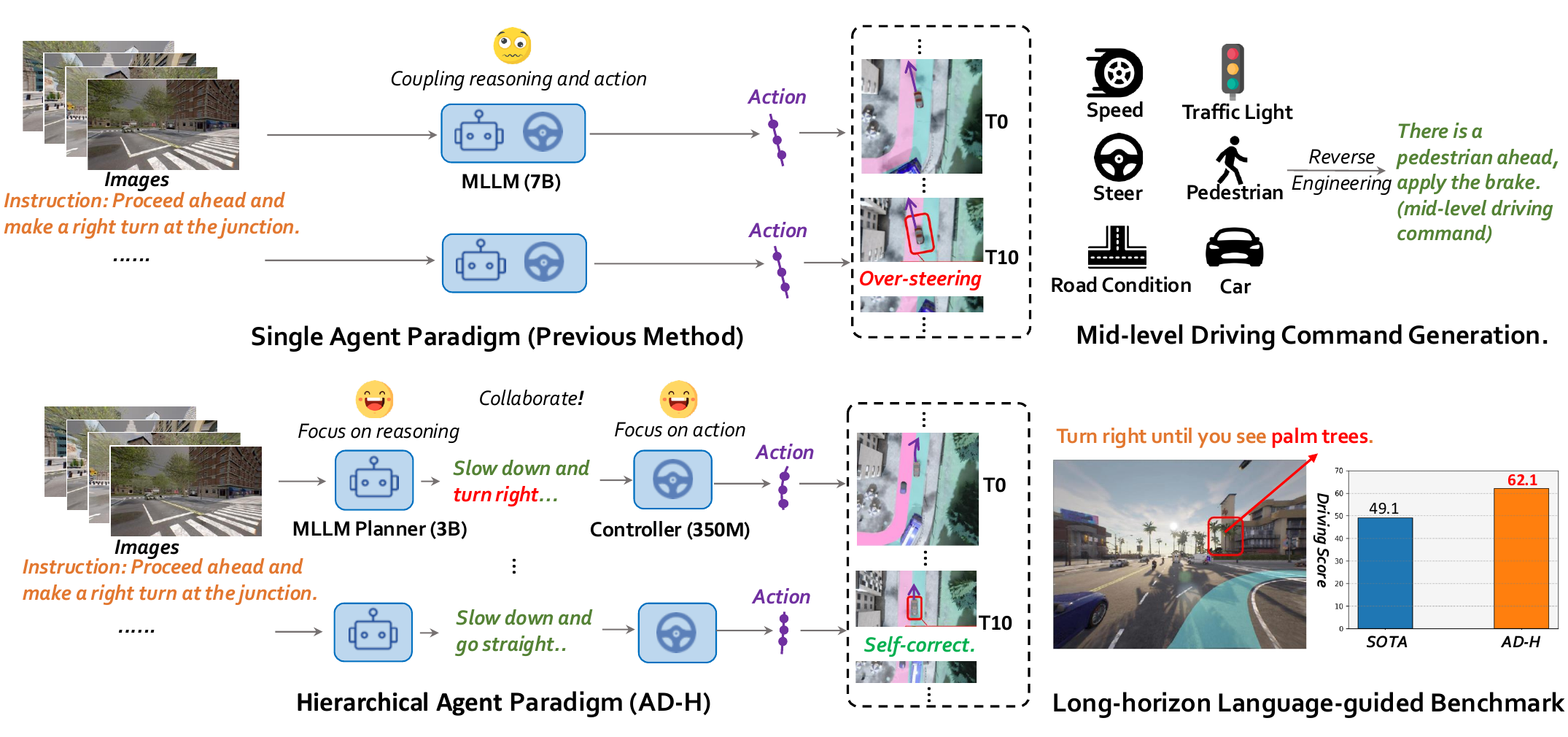}
\caption{The paradigm of language-guided hierarchical autonomous driving. Instead of using an end-to-end MLLM to map high-level instructions directly to low-level actions, which often leads to poor generalization, we adopt a hierarchical design where the MLLM generates mid-level language commands and a lightweight controller translates them into actions.}
\label{fig:teaser}
\end{figure*}

Language-guided autonomous driving—where vehicles interpret and execute natural language instructions—has emerged as a key paradigm for building human-centric and trustworthy intelligent transportation systems. With the rapid advancement of Multimodal Large Language Models~(MLLMs)~\cite{liu2024llava, bai2025qwen2}, recent studies~\cite{sima2023drivelm, wang2023drivemlm, chen2023drivingwithllm, liu2023mtd, sha2023languagempc, wen2023dilu, tian2024drivevlm, mei2024leapad, winter2025bevdriver} have explored using MLLMs as the central agent for perception, reasoning, and interaction, achieving promising results. However, most existing language-guided systems adopt a \emph{single} MLLM agent that directly maps high-level instructions to low-level control signals~\cite{shao2023lmdrive, winter2025bevdriver}, which is misaligned with the pre-training objective of natural language generation. Consequently, precise control relies heavily on domain-specific fine-tuning and tends to overfit specific scenarios and instruction patterns. As illustrated in Figure~\ref{fig:teaser}, such models may fail in unseen situations (e.g., oversteering cases), motivating a central question: \textit{Can we design an autonomous driving system that better leverages the emergent capabilities of pre-trained MLLMs for robust reasoning and generalization to unfamiliar instructions and scenarios?}

To answer the above question, we explore the idea of hierarchical policy~\cite{belkhale2024rt-h,chen2024rh20t}. Rather than directly predicting the final control signals, we propose to fill the gap between high-level language instructions and low-level control signals with intermediate mid-level driving commands. 
This structure offers two key benefits:
First, mid-level commands provide finer granularity than abstract high-level language instructions and are closer to the control space, enabling better responsiveness to real-time feedback.
Second, they are still language-aligned and thus well-suited for MLLMs, allowing the system to better utilize pre-trained world knowledge. Moreover, this decomposition facilitates more flexible human interaction, shared policy learning, and transferability across tasks, promoting generalization~\cite{belkhale2024rt-h}.

In light of the above motivation, we design a Hierarchical Multi-Agent System for Autonomous Driving (AD-H), which comprises two agents: a MLLM-based planner and a lightweight controller. As shown in Figure~\ref{fig:teaser}, the planner aims to perform planning and decision-making based on the input contextual high-level instruction and predicts a mid-level command at each decision frame. The mid-level command is then decoded into the low-level control signals by the controller given the current visual input and the contextual instruction. The high-level planner and low-level controller together form a hierarchical policy system, which effectively frees the MLLM from low-level decoding and unlocks its potential for high-level perception, reasoning, and planning. To generate reliable, high-quality mid-level driving commands, we propose a reverse-engineering approach. By leveraging vehicle driving parameters such as speed, steering angle, road conditions, and pedestrian presence, we construct accurate mid-level driving commands using a rule-based method. This approach is highly scalable, allowing us to generate a dataset of 1.15 million frames with hierarchical annotations that across different levels of signals.

Through intensive evaluations under the closed-loop environment of CARLA~\cite{carla}, we show that our AD-H enjoys the following two advantages.  
\textbf{\textit{First, AD-H significantly outperforms 7B-scale models with a much smaller parameter footprint (3B + 350M).}}
Thanks to its hierarchical design, AD-H enables clear task separation between agents: the MLLM-based planner focuses on comprehending human instructions and generating coherent mid-level driving commands, while the lightweight controller is optimized for accurately executing those commands. This division of labor not only enhances overall performance but also improves efficiency. As a result, AD-H achieves superior results on the LangAuto benchmark using nearly half the parameter size compared to prior 7B-scale counterparts.
\textbf{\textit{Second, AD-H generalizes better to novel and challenging driving scenarios.}}
Our experiments reveal that AD-H exhibits strong emergent generalization capabilities when faced with previously unseen driving situations and corner cases.
For instance, in oversteering scenarios, the planner can issue corrective mid-level commands to guide the vehicle back on track~(Figure~\ref{fig:teaser}). In contrast, existing methods often overfit to the control signal patterns present in training data, resulting in persistent straight-line driving failures~(Figure~\ref{fig:teaser}~(a)).
\textbf{\textit{Third, AD-H demonstrates robust generalization to novel long-horizon instructions.}}
Long-horizon instructions refer to commands that unfold over longer time spans, requiring the model to maintain consistent reasoning and planning across extended temporal sequences.
To evaluate this, we introduce a long-horizon benchmark featuring complex, long-horizon instructions that do not appear in the training set. AD-H successfully understands these instructions, performs coherent high-level planning, and generates accurate mid-level commands at the appropriate decision frames. This leads to substantial improvements in long-horizon driving tasks. In contrast, existing approaches struggle with such tasks, frequently generating incorrect routes.

The contribution of this paper can be summarized as follows:
\begin{itemize}
    \item We propose AD-H, a hierarchical system for autonomous driving, which can significantly unleash the power of MLLMs to achieve higher control precision and generalization.
    \item We construct an autonomous driving dataset with 1.15 million multi-level driving command annotations, which can effectively facilitate hierarchical policy learning. 
    \item We perform intensive experiments and demonstrate that our approach can considerably outperform state-of-the-art methods and exhibits stronger generalization to novel scenarios and long-horizon instructions.
\end{itemize}

\section{Related Work}
\label{sec:related_work}

\subsection{End-to-End methods in Autonomous driving}
In autonomous driving, precise perception~\cite{li2022bevformer, yang2023bevformerv2, liu2023bevfusion, philion2020lift, liang2022bevfusion, Qin_2023_ICCV, li2022voxelfusion, jiao2023msmdfusion, yoo20203dcvf, li2022deepfusion, bai2022transfusion, chen2022autoalign, bevdet, li2022bevstereo, Park2022solo, li2023fbbev, zhou2023matrixvt, wang2023distillbev, wang2023frustumformer, wang2024toward, zhang2023sa, ge2023metabev, li2023bevnext} and planning are critical. To tackle the prevalent issue of long-tail distribution in autonomous driving scenarios, several generative network-based World Models have been developed~\cite{wang2023drivedreamer,jia2023adriver,zhao2024drivedreamer, wen2023panacea}. These networks can generate a vast array of realistic urban street scenes. However, in order to control the vehicle, a separate planning model needs to be designed to utilize the perception results. To solve this problem, many end-to-end autonomous driving models have been proposed, including reinforcement learning based~\cite{prakash2021multi,wu2022trajectory,chitta2022transfuser,codevilla2019exploring,cui2022coopernaut} and imitation learning based methods~\cite{xiao2023scaling,hanselmann2022king}. Besides these, UniAD~\cite{Hu_2023_CVPR} addresses the problem of end-to-end autonomous driving by utilizing multiple modules in BEV space.

Since the emergence of multimodal large models, the field of autonomous driving has been continuously exploring the possibility of using such large models in an end-to-end manner to solve this problem. LLM-Driver~\cite{chen2023driving} uses Vector-former to characterize the perception of the environment by autonomous driving in vector space. Drivegpt4~\cite{xu2023drivegpt4} proposes a novel two-stage training multimodal autonomous driving paradigm, which directly regresses control signals and text responses through multi-frame image input and text instructions. DOLPHINS~\cite{ma2023dolphins} innovatively introduces in-context learning into the autonomous driving framework, which can better mimic human higher-order control abilities. Unlike the methods mentioned above that are trained and tested on static datasets, LMDrive~\cite{shao2023lmdrive} first conducts closed-loop autonomous driving training and testing on the CARLA simulator, demonstrating strong closed-loop control capabilities and scene generalization. As well as several other notable contributions in this area~\cite{li2024driving,zhou2023vision,ding2024holistic,wang2023empowering,ye2024lord,peng2024lc,paul2024lego,wang2024drivecot}. Besides, there have been some exploratory endeavors to leverage agent-based approaches in the domain of autonomous vehicular navigation~\cite{yang2024driving,mao2023language}.

\subsection{Multimodal Large Language Models}
Multimodal Large Language Models (MLLMs) have garnered considerable attention for their remarkable abilities in multimodal perception. Several studies~\cite{liu2024llava, dai2024instructblip, llamaadapter, zhu2023minigpt, lai2023lisa, peng2023kosmos2} focus on integrating visual content into language models, specifically designed to comprehend and reason about images. Among these, LLaVA~\cite{liu2024llava} employs a two-stage instruction-tuning pipeline for comprehensive visual and language understanding. InstructBLIP~\cite{dai2024instructblip} combines the language model with an instruction-aware Q-Former to extract visual content highly pertinent to the provided instruction. Additionally, research~\cite{deshmukh2023pengi, li2023videochat, zhang2023videollama, guo2023point, hong20233d} is expanding MLLMs to include audio, video, and point clouds, enhancing their ability to handle complex multimodal tasks. This integration allows MLLMs to process spatial, auditory, and visual data simultaneously, significantly improving performance in applications like autonomous navigation and multimedia analysis. 

\subsection{LLMs in Task Planning}
In various fields, LLMs have demonstrated their potential in task decomposition for advanced planning. LLMs can incorporate additional visual modules, such as caption descriptions, to perceive environments and influence planning outcomes. SayCan~\cite{ahn2022saycan} integrates LLMs with robotic capabilities, allowing robots to follow complex, long-term natural language instructions. Here, the LLM provides a high-level understanding of the instructions and identifies skills that can offer corresponding low-level controls. To avoid error accumulation due to model stacking, recent research has explored using MLLMs for planning. ViLa~\cite{hu2023ViLa} leverages the world knowledge inherent in MLLMs, including spatial layouts and object attributes, to make more rational task planning for manipulative tasks. RT-H~\cite{belkhale2024rt-h} improves task execution accuracy and learning efficiency by decomposing complex tasks into simple language instructions that are then converted into robotic actions. Nevertheless, it has mainly been investigated under small-scale and static scenarios. It is unknown whether this philosophy can also generalize to large-scale and dynamic autonomous driving environments. More importantly, it lacks suitable training datasets for learning such systems. Our work has filled the above gaps.

\section{Methods}
In this section, we will first delineate the technical details of our proposed AD-H autonomous driving system, and then present the new dataset for training hierarchical multi-agent systems.

\begin{figure*}[ht]
    \centering
    \includegraphics[width=1\textwidth]{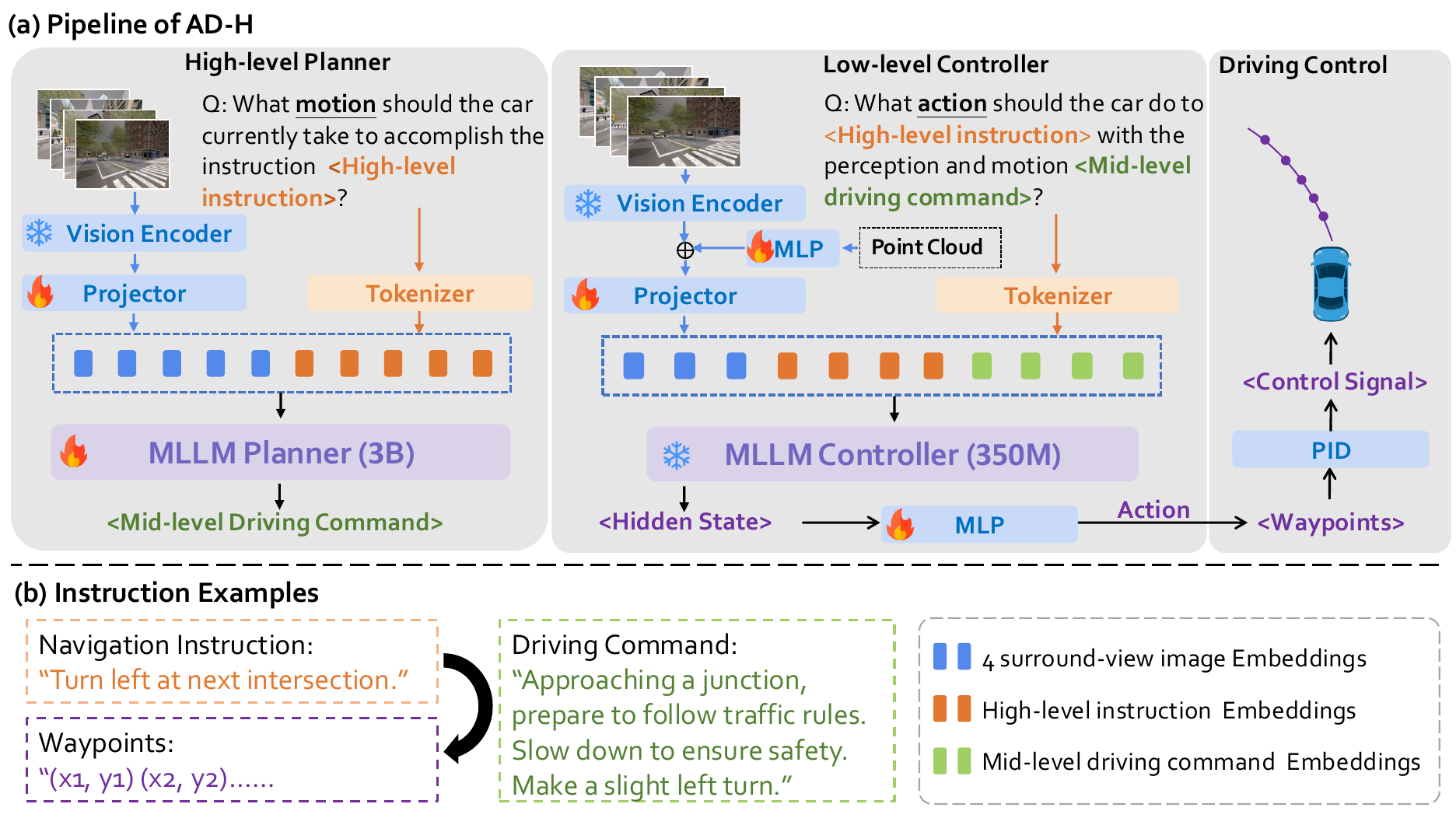}
    \caption{(a) Pipeline of AD-H. The planner breaks down a high-level instruction into mid-level driving commands and the controller decodes low-level waypoints from the mid-level commands. (b) Examples of 
    a high-level instruction, a mid-level command, and low-level waypoints.}
\label{fig:pipeline}
\end{figure*}

\subsection{Overview of AD-H}
The AD-H system consists of two MLLM-based agents, namely a planner and a controller, as illustrated in Figure~\ref{fig:pipeline} (a).
At each decision frame, the planner consumes the current visual input and a high-level contextual instruction (\emph{e.g.}, ``turn left at the next intersection``), performs reasoning \& planning, and makes a decision for the current frame by predicting a mid-level driving command (\emph{e.g.}, ``slow down to ensure safety``). The controller then receives the predicted command and converts it into future waypoints to control the vehicle. The planner and controller, together with the input high-level instruction, the predicted mid-level commands, and low-level waypoints, form a hierarchical structure of action policy for autonomous driving. 
The overall pipeline can be mathematically expressed as  
\begin{equation}\label{eq:overall}
\mathbf{y}_t = g(f(\mathbf{x}_t, \mathbf{i}), \mathbf{x}_t, \mathbf{i}),
\end{equation}
where $\mathbf{i}$ denotes the contextual driving instruction, $\mathbf{x}_t$ and $\mathbf{y}_t$ denotes the visual input and the predicted control signals (\emph{i.e.}, waypoints)  for the $t$-th frame, respectively, and $f$ and $g$ represent the high-level planner and low-level controller, respectively.

\subsection{High-level Planner}
In the AD-H system, the planner focuses solely on high-level decision-making without getting involved in the generation of low-level control signals and therefore becomes more specialized. To do so, the planner needs to perform not only visual perception to understand the surrounding environment as well as its ego status but also effective reasoning and planning to break down the contextual instruction into mid-level driving commands. To this end, we adopt a MLLM as the high-level planner to leverage their strong emergent capabilities (We mainly explore LLaVA-7B~\cite{liu2024llava} and Mipha-3B~\cite{mipha} in our experiments). Figure~\ref{fig:pipeline} (a) illustrates an overview of the MLLM-based planner.  At each decision frame, 4 surround-view images are concatenated and fed into a pre-trained vision encoder~\cite{radford2021learning}. The encoded visual features are further transformed into the textual token space through a projector. Finally, the visual feature together with the tokenized high-level instruction are sequentially fed into the MLLM to predict the mid-level command in an auto-regressive manner.

Through internet-scale pre-training and massive instruction tuning, MLLMs have acquired powerful reasoning ability, along with a wealth of world knowledge, which allows MLLMs to generalize better across various tasks and application scenarios. We then proceed with downstream fine-tuning on our collected autonomous driving dataset (Section~\ref{sec:data}) to teach MLLMs how to generate precise mid-level commands through the next token prediction given the contextual information. Since the driving commands are also natural languages, this downstream task is essentially consistent with the pre-training objectives of MLLMs. As such, the emergent capabilities of the pre-trained MLLMs can be fully unleashed. Our experiments show that the MLLM-based planner can better generalize to novel driving scenarios, long-horizon instructions, as well as unseen environments, and even exhibits self-correction abilities.

\subsection{Low-level
Controller}
The role of the controller is to translate the mid-level driving commands generated by the planner into executable control signals, which is much easier than directly predicting the control signals from the high-level instructions. Therefore, instead of using the 7B LLaVA model~\cite{liu2024llava} as in LMDrive~\cite{shao2023lmdrive}, we adapt the more lightweight OPT-350M~\cite{zhang2022opt} for this purpose. Since OPT-350M is a pure language model, we empower it with visual perception ability by adding an additional vision encoder~\cite{he2016deep} and a Q-Former~\cite{li2023blip2}. 
As shown in Figure~\ref{fig:pipeline}, the pipeline of the controller is similar to that of the planner. The input images are also encoded by the vision encoder and then concatenated with the point cloud features as the same with LMDrive~\cite{shao2023lmdrive}. The concatenated features are projected into the space of textual features through the pre-trained Q-Former. OPT-350M then receives the visual embeddings as well as the textual tokens of the high-level instructions and mid-level commands. The hidden state of its output layer serves as the action embedding and is finally decoded into 5 future waypoints through 2-layer MLP. These waypoints can be input into downstream control algorithms (\emph{e.g.}, PID) to produce numerical information for vehicle control, such as speed, throttle, and steering angle. 
The above pipeline for the controller can be mathematically expressed as   
\begin{equation}\label{eq:controller}
\mathbf{h}_t = g_l(\mathbf{x}_t, \mathbf{i}, \mathbf{c}_t),
\end{equation}
\begin{equation}
\mathbf{y}_t = g_{w}(\mathbf{h}_t),
\end{equation}
where $\mathbf{c}_t$ represents the mid-level command generated by the planner,
$g_l$ and $g_w$ denote the OPT-350M model and the MLP for waypoint regression, respectively, and $\mathbf{h}_t$ indicates the hidden state output of OPT-350M. During training, we feed the ground-truth mid-level command into the controller and minimize the $L_1$ loss between the predicted and ground-truth waypoints.

\section{Training Dataset Construction}\label{sec:data}

Mid-level driving commands provide the core supervision for both the planner and the controller in our hierarchical framework. Existing language-guided autonomous driving datasets~\cite{sima2023drivelm, shao2023lmdrive, zhou2024embodiedelm} mainly contain paired images, trajectories, and high-level navigation instructions collected from real-world logs or closed-loop simulators such as CARLA~\cite{carla}. However, they typically lack explicit, fine-grained mid-level commands, which makes them suboptimal for training hierarchical language–action systems. 

Directly annotating mid-level language commands is also non-trivial: the description must be temporally consistent with the underlying trajectory and must precisely reflect the driver’s intentions (e.g., subtle speed adjustments or slight steering corrections). To avoid expensive and noisy manual labeling, we adopt a reverse-engineering strategy that converts existing trajectories into structured, text-based mid-level commands via a rule-based engine. Concretely, for each frame, we first extract a set of interpretable driving signals, including: (1) ego-vehicle state (speed, steering angle, brake status, target speed), and (2) perception-related factors (road topology and junctions, traffic lights and signs, as well as nearby vehicles, bikes, and pedestrians). These signals are then fed into a collection of hand-crafted rules that map them to a small set of atomic sub-commands along four dimensions: \emph{Perception}, \emph{Speed}, \emph{Steer}, and \emph{Brake}, as shown in Table~\ref{tab:sub_command}. The final mid-level command is obtained by composing these sub-commands into a concise natural-language instruction.

Algorithm~\ref{alg:mid_level_command} illustrates the rule-based generation process for the \emph{Speed} and \emph{Steer} dimensions using the ego speed, target speed, steering angle, and brake signal as inputs. Similar rule templates are defined for perception-related events (e.g., vehicles at a junction, pedestrians ahead, traffic light status) and for braking behavior. The complete rule set and additional implementation details are provided in the supplementary material.

\begin{algorithm}[h]
\caption{Generate Mid-Level Driving Sub-Commands (Speed \& Steer)}
\label{alg:mid_level_command}
\begin{algorithmic}[1]
\STATE \textbf{Input:} $speed, target\_speed, steer, brake$
\STATE \textbf{Output:} $speed\_cmd, steer\_cmd, brake\_cmd$
\vspace{0.3em}
\IF{$brake$ or $should\_brake$}
    \STATE $brake\_cmd \leftarrow$ ``Apply brakes safely.''
\ELSE
    \STATE $brake\_cmd \leftarrow$ \textbf{None}
\ENDIF
\vspace{0.3em}
\IF{$speed < target\_speed$}
    \IF{$speed = 0$}
        \STATE $speed\_cmd \leftarrow$ ``Start accelerating gradually towards the target speed.''
    \ELSIF{$target\_speed - speed > threshold$}
        \STATE $speed\_cmd \leftarrow$ ``Significantly below target speed, accelerate if safe.''
    \ELSE
        \STATE $speed\_cmd \leftarrow$ ``Slightly below target speed, gently increase acceleration.''
    \ENDIF
\ELSIF{$speed > target\_speed$}
    \STATE $speed\_cmd \leftarrow$ ``Above target speed, decelerate.''
\ELSE
    \STATE $speed\_cmd \leftarrow$ ``Maintain current speed to match the target speed.''
\ENDIF
\vspace{0.3em}
\IF{$steer > angle\_threshold$}
    \STATE $steer\_cmd \leftarrow$ ``Make a slight right turn.'' \COMMENT{Or ``Steer right sharply'' if $|steer|$ is large}
\ELSIF{$steer < -angle\_threshold$}
    \STATE $steer\_cmd \leftarrow$ ``Make a slight left turn.'' \COMMENT{Or ``Steer left sharply'' if $|steer|$ is large}
\ELSE
    \STATE $steer\_cmd \leftarrow$ ``Keep the steering wheel straight.''
\ENDIF
\end{algorithmic}
\end{algorithm}

\begin{figure*}[ht]
    \centering
    \includegraphics[width=1\textwidth]{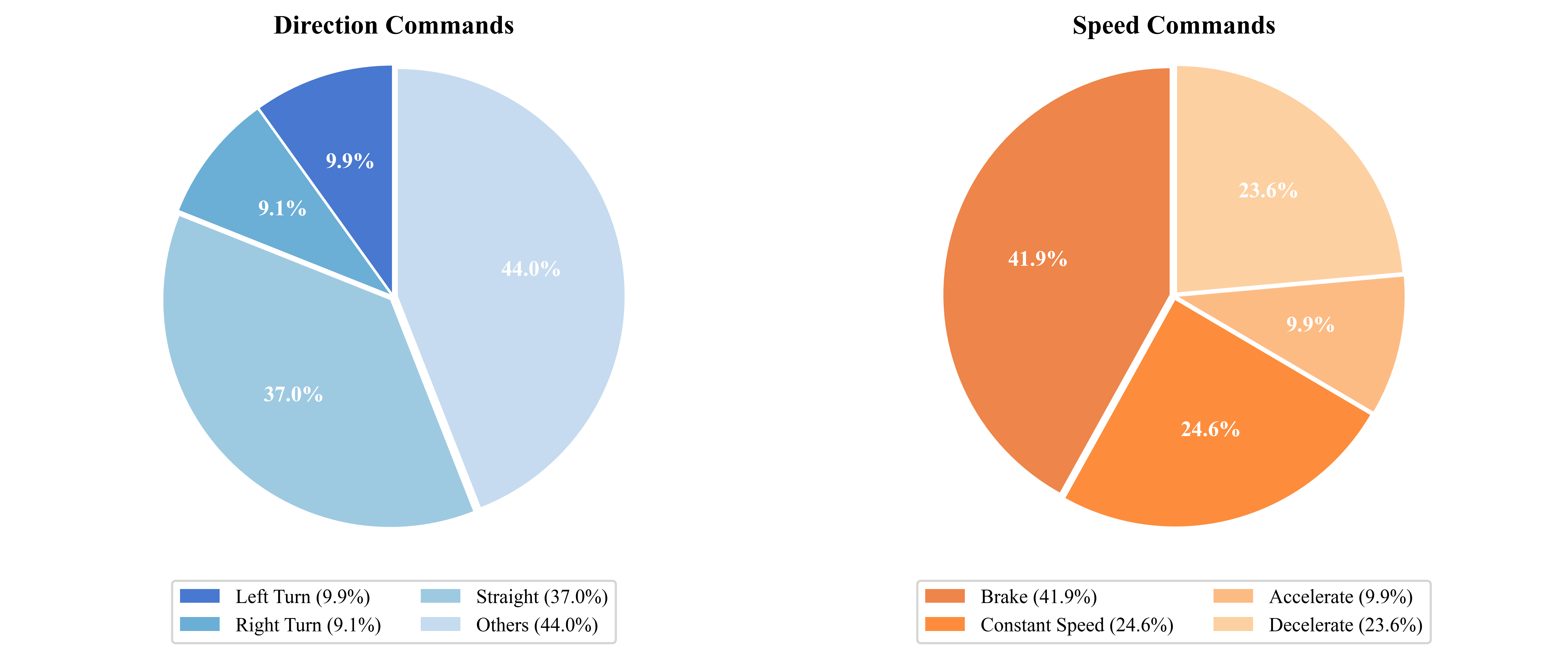}
    \caption{Distribution of direction-related and speed-related sub-commands in the AD-H dataset, illustrating the diversity of generated mid-level driving behaviors.}
    \label{fig:data_vis}
\end{figure*}

Using the LMDrive dataset~\cite{shao2023lmdrive} as the trajectory source, we apply the above rule-based generator to construct our AD-H dataset, which contains 1.15 million frames. Each frame is associated with multi-modal inputs (RGB images and LiDAR point clouds), a high-level navigation instruction, the generated mid-level command, and the corresponding low-level control actions. Figure~\ref{fig:data_vis} visualizes the distribution of direction-related and speed-related sub-commands, showing that our rule design induces a diverse set of motion behaviors that cover a wide range of driving patterns.

Table~\ref{tab:sub_command} summarizes the complete set of 26 atomic driving sub-commands used in AD-H. They are grouped into four categories: perception-aware descriptions of the surrounding scene, speed-related adjustments, steering behaviors, and braking actions. By combining these atomic sub-commands, our rule-based generator can produce more than 170 distinct mid-level driving commands, providing rich and interpretable supervision for training the hierarchical planner and controller.

\begin{table}[h]
\centering
\caption{Full list of the 26 atomic driving sub-commands in the AD-H dataset. Different combinations of these sub-commands yield over 170 distinct mid-level driving commands.}
\label{tab:sub_command}
\begin{tabular}{l|p{6.5cm}}
\toprule
\textbf{Type} & \textbf{Driving command} \\
\midrule

\multirow{13}{*}{\textbf{Perception}} 
 & Approaching a junction, prepare to follow traffic rules. \\
 & A vehicle is present at the junction. Be cautious. \\
 & Multiple vehicles are present at the junction. Be cautious. \\
 & Watch out for the car ahead, there's a vehicle in front. \\
 & Watch out for the cars ahead, there are multiple vehicles in front. \\
 & A vehicle is present in the lane. Be cautious. \\
 & Multiple vehicles are present in the lane. Be cautious. \\
 & There is a bike ahead. Be cautious. \\
 & Multiple bikes are ahead. Be cautious. \\
 & There is a pedestrian ahead. Be cautious. \\
 & Multiple pedestrians are ahead. Be cautious. \\
 & There is a red light ahead. \\
 & There is a stop sign ahead. \\
\midrule

\multirow{7}{*}{\textbf{Speed}} 
 & Slow down to ensure safety. \\
 & Start accelerating gradually towards the target speed. \\
 & Remain stopped due to brake application. \\
 & Significantly below target speed, accelerate if safe. \\
 & Slightly below target speed, gently increase acceleration. \\
 & Above target speed, decelerate. \\
 & Maintain current speed to match the target speed. \\
\midrule

\multirow{5}{*}{\textbf{Steer}} 
 & Steer right sharply. \\
 & Make a slight right turn. \\
 & Steer left sharply. \\
 & Make a slight left turn. \\
 & Keep the steering wheel straight. \\
\midrule

\textbf{Brake} & Apply brakes safely. \\
\bottomrule
\end{tabular}
\end{table}

\section{Experiment}
\label{sec:experiment}

\subsection{Experiment Settings}
\label{sec:experiment_details}
\subsubsection{Implementation Details}
\paragraph{Training Details}
The MLLM planner is the pre-trained LLaVA-7B-V1.5~\cite{liu2024llava} and Mipha-3B~\cite{mipha}. The MLLM controller is the OPT-350M~\cite{zhang2022opt} with a ResNet50 vision encoder~\cite{he2016deep}.
We fine-tune the MLLM planner with full parameters. 
The initial learning rate is set to 2e-5, and a few steps of warm-up are incorporated into the training process. For the MLLM controller, the training setup is the same with LMDrive. The learning rate is set to 1e-5. Training is conducted using the Adam optimizer on 4 NVIDIA A800 GPUs. More details are presented in supplementary materials.

\paragraph{Inference Details}
For inference, the system utilizes two computers, each equipped with a single NVIDIA RTX 3090 GPU. The CARLA simulator and the low-level controller share one machine, with a combined memory usage of approximately 12GB (7GB for CARLA and 5GB for the controller). These components communicate via TCP/IP to ensure smooth model swapping without restarting the CARLA simulator. The high-level planner is deployed on a separate machine with similar GPU specifications, also using around 12GB of memory, and communicates with both the CARLA simulator and low-level controller via TCP/IP. Additionally, a quantized version of the planner (supporting 4-bit and 8-bit quantization) is available to enable efficient inference on a single RTX 3090, further reducing computational costs.

\subsubsection{Evaluation Benchmarks and Metrics}
We conduct standard closed-loop evaluations using CARLA simulator~\cite{carla} on the LangAuto Benchmark~\cite{shao2023lmdrive}. On top of LangAuto, we further build two additional benchmarks termed LangAuto-Long-Horizon and LangAuto-Novel-Environment, which contain long-horizon instructions and novel environments, respectively. We present their details as follows.

\paragraph{LangAuto Benchmark.} 
The LangAuto benchmark encompasses a variety of test routes spanning eight towns, diverse weather conditions, and misleading interference. Throughout the testing procedure, algorithms navigate vehicles within the environment, utilizing solely language commands and visual input. The LangAuto benchmark is further divided into three sub-tracks: LangAuto, with routes longer than 500 meters; LangAuto-Short, with routes between 150 and 500 meters; and LangAuto-Tiny, with routes shorter than 150 meters. We follow the prior method~\cite{shao2023lmdrive} and perform evaluations separately on these three sub-tracks.

\paragraph{LangAuto-Long-Horizon Benchmark.}
Planning and decision-making over long time horizons is a central challenge in embodied AI~\cite{pirk2020longhorizon, huang2022language, zeng2022socratic, ahn2022can, huang2022inner}, where an agent must follow a sequence of sub-instructions to accomplish a single high-level goal. To evaluate AD-H in such settings, we construct the LangAuto-Long-Horizon benchmark by extending the LangAuto-Tiny benchmark. Specifically, we merge multiple short navigation instructions that originally appear sequentially within the same episode into a single long-horizon instruction that summarizes the entire route. For example, the instruction series ``Alright, you can start driving'', ``Keep on rolling straight till you get to the next junction,'' and ``Continue in a straight line along your current path'' are consolidated into one directive: ``Go straight ahead, turn left at the end of the road, then continue straight.''

Since neither AD-H nor the baseline LMDrive model maintains explicit historical frame information or instruction history, a long-horizon instruction may become ambiguous if it does not clearly indicate where a turn should be executed. To mitigate this, we incorporate salient environmental cues into the long-horizon instructions, so that the correct turning point can be unambiguously identified from a single frame. For instance, we use descriptions such as ``Go straight until you see a turning point with palm trees ahead, then turn right and follow the road.''

All long-horizon instructions in LangAuto-Long-Horizon are strictly held out from the LMDrive-H training set. Consequently, this benchmark not only tests the ability of autonomous driving systems to handle long-horizon language guidance, but also assesses their generalization to previously unseen navigation instructions. Representative examples are shown in Table~\ref{table:driving_commands}, and the full set of long-horizon instructions is provided in the Supplementary Material.

\begin{table*}[h]
\centering
\begin{tabular}{l p{16cm}}
\toprule
\textbf{ID} & \textbf{Driving command} \\
\midrule
0  & Go straight ahead, turn left at the end of the road, then continue straight. \\
10 & Go straight until the intersection ahead, then turn right, and continue along the road. \\
12 & Go straight to the first intersection ahead and turn left, then continue straight. \\
20 & Turn right ahead and then go straight. \\
26 & Turn right ahead, go straight, then turn right again. \\
34 & Go straight to the T-junction ahead, then turn left and follow the route. \\
44 & Go straight to a crossroads, then turn left, then continue straight. \\
46 & Go straight to the T-junction, turn right, and continue straight. \\
48 & Follow the route, and continue straight when you reach the crossroads. \\
57 & Go straight to the intersection where, on the left front side, there is an open space with some parked vehicles, and turn left. \\
68 & Keep going along this road. \\
70 & Turn left at the T-junction ahead, then follow the road. \\
74 & Turn left ahead when you reach the cornfield, then turn left again when you encounter an open area. \\
81 & Slightly turn left along the road ahead, then turn right, turn left at the T-junction, and then go straight. \\
84 & Go straight until you see a turning point with palm trees ahead, then turn right and follow the road. \\
88 & Turn right at the T-junction, go straight, then turn right at the T-junction where there are grid lines on the ground. Then continue straight. \\
\bottomrule
\end{tabular}
\vspace{0.5mm}
\caption{Representative long-horizon navigation instructions in the LangAuto-Long-Horizon benchmark, used as mid-level guidance for the controller.}
\label{table:driving_commands}
\end{table*}

\paragraph{LangAuto-Novel-Environment Benchmark.}
To rigorously assess the generalization ability of language-guided autonomous driving systems under unseen environments, we construct the LangAuto-Novel-Environment benchmark. Specifically, we follow a train-in-one-town, test-in-unseen-towns protocol:
all models are retrained from scratch on Town01, and their performance is evaluated on the remaining seven towns (Town02–Town07 and Town10), which are entirely unseen during training.
To ensure strict non-overlap, we additionally remove from the LMDrive-H training set all data associated with these seven towns. This setup enforces a clean distribution shift and allows us to measure the true generalization capability of each method when deployed in novel environments.

\paragraph{Evaluation Metrics.} We employ three widely used evaluation metrics from the CARLA LeaderBoard~\cite{carla}, including route completion (RC), infraction score (IS), and driving score (DS). Among them, RC measures the percentage of the planned route that is successfully completed, with a specific focus on the distance covered along designated segments. Any significant deviation from the intended route leads to the episode being marked as a failure. The IS metric keeps track of violations such as collisions or traffic infractions, which decrease the score with each offense. The DS metric combines both the RC and IS scores to provide a comprehensive assessment of progress and safety, serving as the primary evaluation metric.

\subsection{Results and Analysis}

In this section, we mainly investigate the performance of the autonomous driving models from four perspectives: (1) standard evaluation in a closed-loop manner, (2) generalization to novel long-horizon instructions, (3) generalization to novel environments, and (4) performance achieved by using different MLLMs as planners.
As the LangAuto is a new benchmark, only the result of LMDrive~\cite{shao2023lmdrive} is available. Therefore, we adopt LMDrive as our main competitor. It should be noted that 
LMDrive is one of the pioneering works in language-guided closed-loop driving and can serve as a strong baseline of our method without using hierarchical agents.

\subsubsection{Closed-loop Driving Performance}
\label{sec:exp_performance}

Table~\ref{tab:exp-performance} reports the quantitative comparisons on the LangAuto benchmark. It shows that our AD-H significantly outperforms all LMDrive variants across the three sub-tracks, especially in terms of the main score DS. On the full LangAuto benchmark, AD-H~(LLaVA-7B + OPT-350M) achieves a DS of \textbf{44.0}, improving over the best LMDrive baseline (36.2) by \textbf{+7.8 points} (\textbf{+21.5\%}). The RC score is similarly enhanced, increasing from 46.5 to \textbf{53.2} (\textbf{+14.4\%}). Consistent gains are observed in the LangAuto-Short setting, where AD-H improves the DS from 50.6 to \textbf{56.1} (\textbf{+10.9\%}) and the RC from 60.0 to \textbf{68.0} (\textbf{+13.3\%}). The improvement becomes even more pronounced in the LangAuto-Tiny benchmark: AD-H achieves a DS of \textbf{77.5}, outperforming LMDrive (66.5) by \textbf{+11.0 points} (\textbf{+16.5\%}), with RC also rising from 77.9 to \textbf{85.1} (\textbf{+9.2\%}).

\begin{figure}[h]
    \centering
    \includegraphics[width=0.5\textwidth]{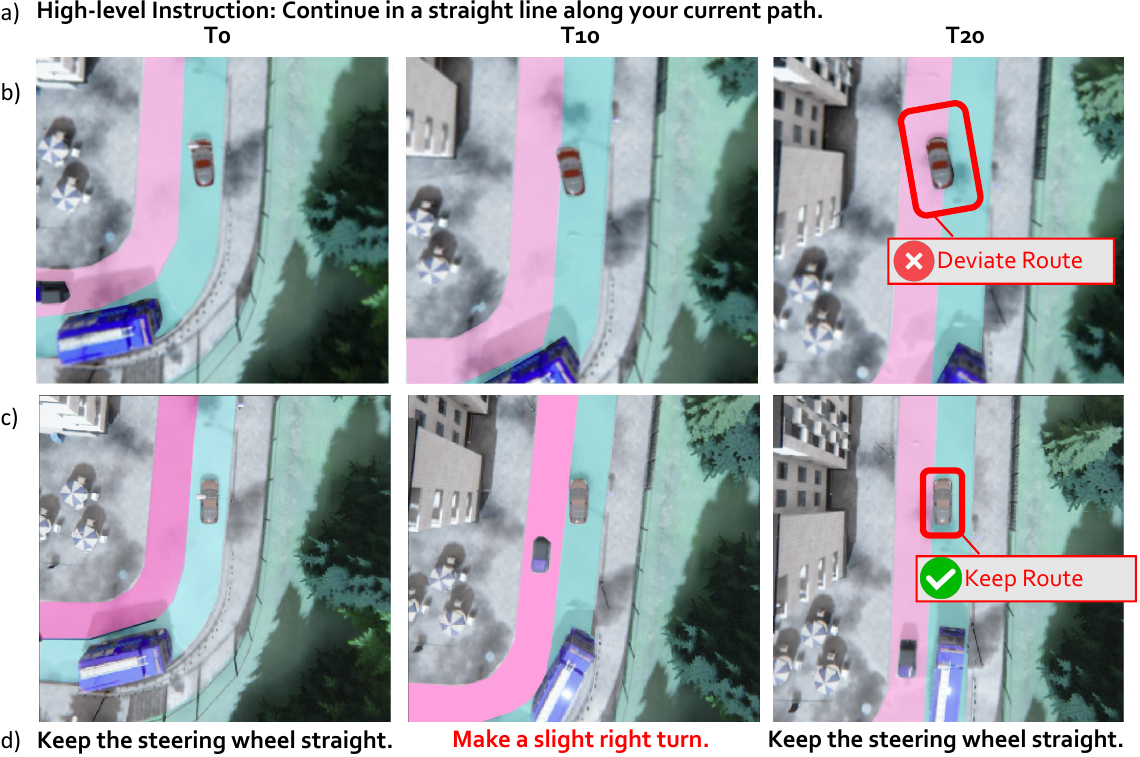}
    \caption{Results of self-correction scenario. (a) High-level instruction; (b) Visualization results of LMDrive; (c) Visualization results of AD-H; (d) Mid-level driving commands predicted by the planner of AD-H. The visual results show that LMDrive maintains a straight trajectory after oversteering, deviating from the intended path. However, AD-H is able to issue precise commands to guide the vehicle back on track.
    }
\label{fig:visual_0}
\end{figure}

Moreover, we find that even the smaller AD-H configuration (Mipha-3B + OPT-350M) performs consistently better than all 7B LMDrive models (e.g., 41.1 vs.\ 36.2 on LangAuto), further validating the effectiveness of the hierarchical paradigm—its performance gain is rooted in the decomposition of planning and control rather than in model size. Through extensive analysis, we further observe that AD-H exhibits frequent self-correction behaviors. As shown in Figure~\ref{fig:visual_0}, LMDrive fails to recognize the road conditions after a left turn and continues executing the previously received high-level command, causing the vehicle to cross the lane boundary. In contrast, AD-H dynamically re-generates mid-level commands based on the current visual context, enabling the controller to adjust its posture accordingly and substantially reducing the risk of lane departures, traffic jams, and accident-prone behaviors.

\begin{table*}[h!]
\centering
\resizebox{1.0\textwidth}{!}{%
\begin{tabular}{@{}lccccccccc@{}}
\toprule
\textbf{Method} & \multicolumn{3}{c}{\textbf{LangAuto}} & \multicolumn{3}{c}{\textbf{LangAuto-Short}} & \multicolumn{3}{c}{\textbf{LangAuto-Tiny}} \\ 
\cmidrule(r){2-4} \cmidrule(lr){5-7} \cmidrule(l){8-10}
& \textbf{DS($\uparrow$)} & \textbf{RC($\uparrow$)} & \textbf{IS($\uparrow$)} & \textbf{DS($\uparrow$)} & \textbf{RC($\uparrow$)} & \textbf{IS($\uparrow$)} & \textbf{DS($\uparrow$)} & \textbf{RC($\uparrow$)} & \textbf{IS($\uparrow$)} \\
\midrule
Random Init& 10.7 & 16.2 & 0.63 & 14.2 & 20.1 & 0.72 & 20.1 & 24.7 & 0.75 \\
\midrule
LMDrive~(Llama-7B)~\cite{shao2023lmdrive}& 31.3 & 37.1 & 0.82 & 42.8 &  49.1 & 0.87 & 52.5 &  57.8 & \textbf{0.91} \\
LMDrive~(Llama2-7B)~\cite{shao2023lmdrive}& 32.8 & 40.1 & 0.81 & 44.8 &  53.5 & 0.84 & 56.1 &  64.2 & 0.87 \\
LMDrive~(Vicuna-7B)~\cite{shao2023lmdrive}& 33.5 & 39.3 & 0.83 & 45.3 & 54.3 & 0.83 & 55.5 & 63.1 & 0.88 \\
LMDrive~(Vicuna-v1.5-7B)~\cite{shao2023lmdrive}& 34.0 & 39.0 & 0.85 & 47.0 & 56.5 & 0.83 & 59.0 & 69.9 & 0.84 \\
LMDrive~(LLaVA-7B)~\cite{shao2023lmdrive}& 36.2 & 46.5 & 0.81 & 50.6 & 60.0 & 0.84 & 66.5 & 77.9 & 0.85 \\
BEVDriver~(Llama3.1-8B-I)~\cite{winter2025bevdriver}& 36.2 & 46.5 & 0.81 & 50.6 & 60.0 & 0.84 & 66.5 & 77.9 & 0.85 \\
\midrule
AD-H~(Mipha-3B + OPT-350M)&41.1 & 48.5 & \textbf{0.86} & 54.3 & 61.8 & \textbf{0.86} & 68.0 & 74.4 & 0.87 \\
AD-H~(LLaVA-7B + OPT-350M)&\textbf{44.0} & \textbf{53.2} & 0.83 & \textbf{56.1} & \textbf{68.0} & 0.78 & \textbf{77.5} & \textbf{85.1} & \textbf{0.91} \\
\bottomrule
\end{tabular}%
}
\vspace{1mm}
\caption{Comparison on the \textsc{LangAuto} benchmark. 
The best results are highlighted in \textbf{bold}. 
$\uparrow$ indicates that higher values are better, while $\downarrow$ indicates that lower values are better. 
All models are evaluated under the same settings for fair comparison.}
\label{tab:exp-performance}
\end{table*}

\subsubsection{Generalization to Long-Horizon Instruction}
\begin{figure}[h]
    \centering
    \includegraphics[width=0.5\textwidth]{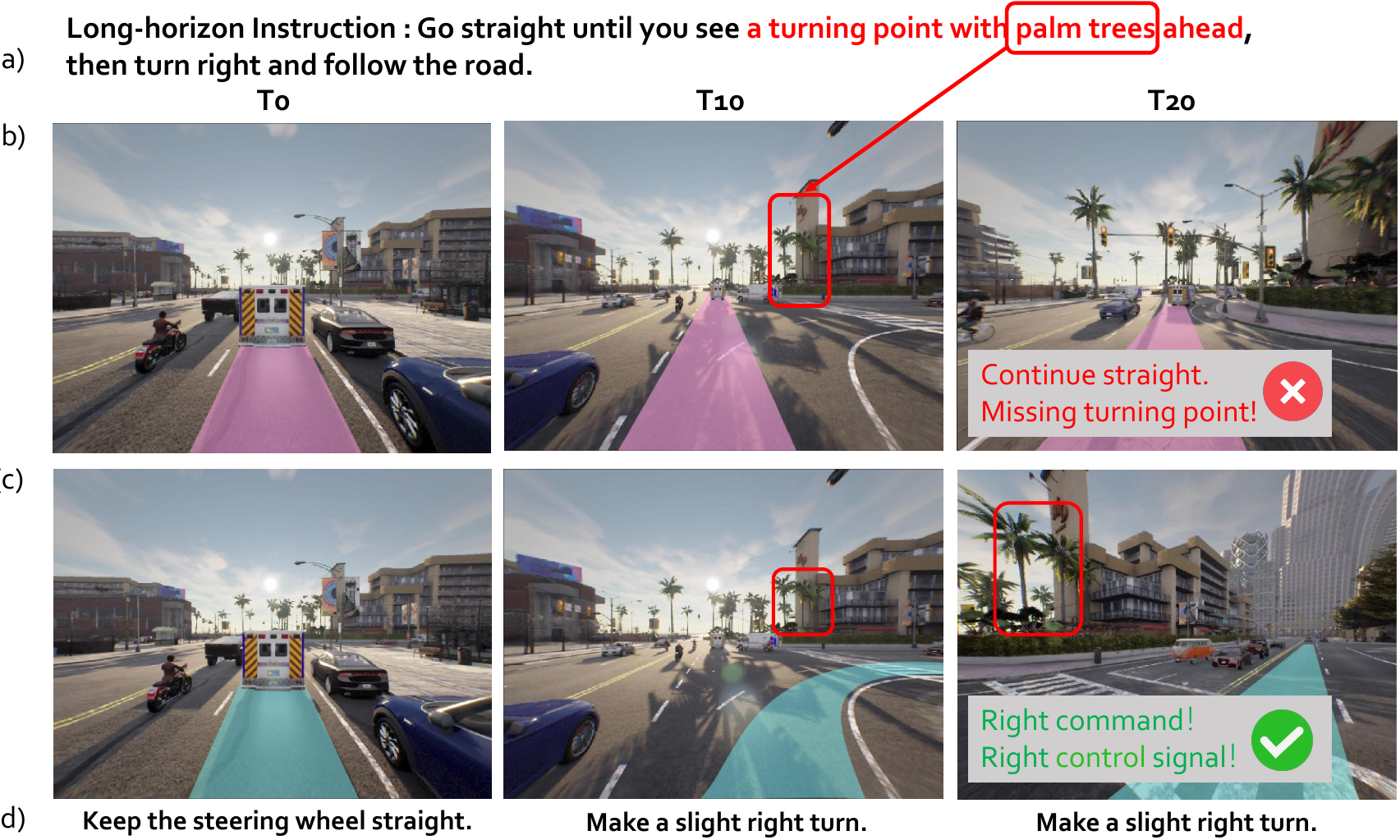 }
    \caption{Results with long-horizon instructions (a). (b) LMDrive persists in following the initial instructions, continuing forward; (c) AD-H can adeptly assess environmental cues to determine the appropriate timing for turning; (d) Mid-level commands produced by AD-H.
    }
\label{fig:long_horizon}
\end{figure}

Table~\ref{table:long-hori} presents the results on the LangAuto-Long-Horizon benchmark, where the high-level navigation instructions are long-range and provided only at the beginning of each task. Since both LMDrive and AD-H are trained exclusively with short-horizon instructions during driving, this benchmark evaluates their ability to generalize to unseen long-horizon instruction settings. Despite the increased difficulty, AD-H still delivers strong performance and surpasses LMDrive by a considerable margin. Quantitatively, AD-H~(LLaVA-7B + OPT-350M) achieves a DS of \textbf{62.1}, outperforming LMDrive (49.1) by \textbf{+13.0 points} (\textbf{+26.5\%}) and improving IS from 0.87 to \textbf{0.88}. Meanwhile, AD-H~(Mipha-3B + OPT-350M) achieves a remarkable RC of \textbf{73.2}, significantly higher than LMDrive’s 56.4 (\textbf{+29.7\%}). As illustrated in Figure~\ref{fig:long_horizon}, LMDrive—directly predicting control signals—fails to fully capture the global instruction context and continues straight instead of making the required right turn. In contrast, our hierarchical AD-H continuously analyzes both the initial instruction and the evolving visual environment, generating precise mid-level commands that guide the controller throughout the trajectory. These results demonstrate the strong generalization capability of AD-H in handling long-horizon, instruction-driven navigation tasks.

\begin{table}[h]
\centering
\begin{tabular}{lccc}
    \toprule
    \textbf{Method} & \textbf{DS($\uparrow$)} & \textbf{IS($\uparrow$)} & \textbf{RC($\uparrow$)} \\
    \midrule
    LMDrive (LLaVA-7B)~\cite{shao2023lmdrive} & 49.1 & 0.87 & 56.4 \\
    AD-H (Mipha-3B + OPT-350M) & 55.1 & 0.74 & \textbf{73.2} \\
    AD-H (LLaVA-7B + OPT-350M) & \textbf{62.1} & \textbf{0.88} & 68.3 \\
    \bottomrule
\end{tabular}
\vspace{0.5mm}
\caption{
Performance on the LangAuto-Long-Horizon benchmark. 
This setting evaluates zero-shot generalization to long-horizon navigation instructions, 
which are only provided at the beginning of the task. 
AD-H consistently outperforms LMDrive, demonstrating stronger capability in interpreting 
and executing long-range instructions. Bold numbers indicate the best results.
}
\label{table:long-hori}
\end{table}

\subsubsection{Generalization to Novel Environments}
\label{sec:exp_generalization}
We further evaluate zero-shot generalization to unseen scenes on the LangAuto-Novel-Environment benchmark. 
As shown in Table~\ref{table:new_env}, both variants of AD-H consistently outperform LMDrive across all metrics, including driving success (DS), infraction score (IS), and route completion (RC). 
Notably, AD-H with the compact Mipha-3B backbone already achieves substantial gains over LMDrive, while AD-H with LLaVA-7B further pushes all metrics to the best levels, indicating that our hierarchical planner–controller architecture enables strong generalization even with smaller MLLMs and scales gracefully with more powerful backbones.

\begin{table}[h]
\centering
\begin{tabular}{lccc}
    \toprule
    \textbf{Method} & \textbf{DS($\uparrow$)} & \textbf{IS($\uparrow$)} & \textbf{RC($\uparrow$)} \\
    \midrule
    LMDrive (LLaVA-7B)~\cite{shao2023lmdrive} & 46.0 & 0.77 & 57.7 \\
    AD-H (Mipha-3B + OPT-350M) & 58.1 & 0.86 & 65.8 \\
    AD-H (LLaVA-7B + OPT-350M) & \textbf{59.9} & \textbf{0.88} & \textbf{67.8} \\
    \bottomrule
\end{tabular}
\vspace{0.5mm}
\caption{
Results on the LangAuto-Novel-Environment benchmark. 
Both AD-H (Mipha-3B) and AD-H (LLaVA-7B) achieve strong zero-shot generalization to unseen environments and clearly surpass LMDrive across all metrics, with the LLaVA-7B variant obtaining the best overall performance (bold). 
}
\label{table:new_env}
\end{table}

\section{Ablation Study}

\paragraph{Ablation of controller.}
We further evaluate controllers of varying sizes, with the results presented in Table~\ref{table:ablation_controller}. Notably, the OPT-350M controller achieves performance comparable to the much larger LLaVA-7B, which can likely be attributed to the high accuracy and granularity of the mid-level commands—effectively reducing the complexity of low-level decoding. In contrast, the performance of OPT-125M is significantly worse, suggesting that overly small controllers struggle to interpret and execute the commands effectively.

\begin{table}[h]
\centering
\begin{tabular}{l l c c c}
    \toprule
    \textbf{Planner} & \textbf{Controller} & \textbf{DS($\uparrow$)} & \textbf{IS($\uparrow$)} & \textbf{RC($\uparrow$)} \\
    \midrule
    LLaVA-7B-v1.5 & LLaVA-7B-v1.5 & 74.6 & 0.80 & \textbf{90.5} \\
    LLaVA-7B-v1.5 & OPT-350M      & \textbf{77.5} & \textbf{0.91} & 85.1 \\
    LLaVA-7B-v1.5 & OPT-125M      & 33.9 & 0.90 & 35.9 \\
    \bottomrule
\end{tabular}
\vspace{0.5mm}
\caption{
Ablation study on controller size with a fixed planner (LLaVA-7B-v1.5).
Bold numbers indicate the best performance for each metric. 
The OPT-350M controller performs comparably to the much larger LLaVA-7B, 
while the OPT-125M controller shows a substantial performance drop.
}
\label{table:ablation_controller}
\end{table}

\paragraph{Ablation of Planner}
We investigate how the accuracy of planner-predicted mid-level driving commands impacts closed-loop performance when using a fixed controller (OPT-350M). 
To quantify planner accuracy, we conduct a human evaluation on all scenarios in the LangAuto-Tiny benchmark: for each scenario, we run the planner in closed loop and randomly sample 100 time steps from its inference trajectories, then manually judge whether the predicted mid-level command correctly reflects the underlying control behavior, considering both steering direction and speed regulation. 
This sampling-and-annotation process is repeated three times with different random seeds, and we report the averaged accuracy as the planner's mid-level command accuracy.

Table~\ref{table:ablation_planner_accuracy} compares two planners with notably different command accuracies. 
A higher mid-level command accuracy correlates with better downstream Driving Score (DS), Rule Compliance (RC), and Intervention Score (IS), indicating that precise mid-level guidance reduces ambiguity for the controller and improves overall driving quality.
The 7.3\% absolute increase in command accuracy (87.0\% $\rightarrow$ 94.3\%) yields notable gains in DS (+9.5), RC (+10.7), and IS (+0.04). 
This suggests that accurate and granular mid-level commands effectively constrain the action space for the controller, reducing misinterpretation and intervention needs. 

\begin{table}[h]
\centering
\begin{tabular}{l l c c c c}
    \toprule
    \textbf{Planner} & \textbf{Controller} & \textbf{Accuracy(\%)} & \textbf{DS($\uparrow$)} & \textbf{RC($\uparrow$)} & \textbf{IS($\uparrow$)} \\
    \midrule
    Mipha-3B   & OPT-350M & 87.0 & 68.0 & 74.4 & 0.87 \\
    LLaVA-7B   & OPT-350M & \textbf{94.3} & \textbf{77.5} & \textbf{85.1} & \textbf{0.91} \\
    \bottomrule
\end{tabular}
\vspace{0.5mm}
\caption{
Ablation on planner mid-level command accuracy with a fixed controller (OPT-350M).
Higher command accuracy (LLaVA-7B: 94.3\%) consistently improves closed-loop performance (DS/RC/IS) compared to a lower-accuracy planner (Mipha-3B: 87.0\%).
}
\label{table:ablation_planner_accuracy}
\end{table}

\paragraph{Ablation of Mid-level Driving Commands.}
We conduct an ablation study to examine the effect of mid-level driving commands on controller performance. Specifically, we retrain the controller using LLaVA-1.5-7B under two input settings: (1) high-level instructions only, and (2) a combination of high-level instructions and mid-level driving commands (referred to as the AD-H setting). The training configuration remains consistent with that described in Section~\ref{sec:experiment_details}.

During inference, setting (1) directly inputs only high-level instructions into the controller without any planner involved. In setting (2), a planner (Mipha-3B) first generates mid-level driving commands, which are then concatenated with the high-level instructions and fed into the controller. The corresponding closed-loop performance on the LangAuto benchmark is reported in Table~\ref{table:ab_mid_level_close}.
The results show that the inclusion of mid-level driving commands significantly improves the overall driving performance, highlighting their effectiveness in guiding low-level control.

\begin{table}[h]
\centering
\begin{tabular}{lccc}
    \toprule
    \textbf{Method} & \textbf{DS($\uparrow$)} & \textbf{IS($\uparrow$)} & \textbf{RC($\uparrow$)} \\
    \midrule
    Instruction Only & 60.7 & \textbf{0.91} & 65.7 \\
    Instruction + Mid-level Command & \textbf{74.6} & 0.80 & \textbf{90.5} \\
    \bottomrule
\end{tabular}
\vspace{0.5mm}
\caption{
Closed-loop ablation on the effect of mid-level driving commands. 
The AD-H setting (Instruction + Mid-level Command) substantially improves overall driving performance. 
Bold numbers indicate the best results.
}
\label{table:ab_mid_level_close}
\end{table}

\paragraph{Ablation of Dataset}
To assess the accuracy of the mid-level driving commands, we invited three licensed and experienced drivers, all of whom hold valid driving certifications, to evaluate a randomly selected set of 100 commands. The results, summarized in Table~\ref{table:human-performance}, show that all three evaluators achieve accuracies exceeding 90\%, validating the correctness and interpretability of our mid-level driving command design.

\begin{table}[h]
\centering
\begin{tabular}{lccc}
    \toprule
    \textbf{Driver} & \textbf{Correct} & \textbf{Wrong} & \textbf{Accuracy} \\
    \midrule
    Driver 1 & 92 & 8 & 92.0\% \\
    Driver 2 & 93 & 7 & 93.0\% \\
    Driver 3 & 91 & 9 & 91.0\% \\
    \midrule
    \textbf{Average} & \textbf{92.0} & \textbf{8.0} & \textbf{92.0\%} \\
    \bottomrule
\end{tabular}
\vspace{0.5mm}
\caption{
Human evaluation of 100 randomly sampled mid-level driving commands.
All three licensed drivers achieve over 91\% accuracy, confirming the reliability of the mid-level command annotations.
}
\label{table:human-performance}
\end{table}

\paragraph{Ablation of Inference Time and Deployment}
\label{sec:inference_time}
We deploy AD-H on two RTX 3090 GPUs in practice, with one GPU running the planner and the other running the controller. 
To make our system easier to reproduce under a single-GPU setting, we further apply 4-bit quantization to the LLaVA-7B planner and keep the OPT-350M controller in full precision. 
On a single RTX 3090 GPU, we run the CARLA simulator, the 4-bit LLaVA-7B planner, and the OPT-350M controller concurrently, and measure the per-step inference latency of each module. 
The results are summarized in Table~\ref{tab:inference_time}.

\begin{table}[h]
\centering
\begin{tabular}{lccc}
\toprule
 & \textbf{Planner} & \textbf{Controller} & \textbf{Total} \\
\midrule
Latency & 1.15 s/inference & 0.035 s/inference & 1.18 s/inference \\
\bottomrule
\end{tabular}
\vspace{0.5mm}
\caption{Inference latency of the 4-bit LLaVA-7B planner and OPT-350M controller running together with CARLA on a single RTX 3090 GPU.}
\label{tab:inference_time}
\end{table}

Although the controller is lightweight, the overall latency is still dominated by the large planner model, and the current implementation is far from real-time. 
Since deployment-oriented optimization is not the main focus of this work, our inference stack does not adopt the latest engineering techniques (e.g., highly optimized serving engines, KV-cache scheduling, or model distillation). 
In future work, we plan to explore fast–slow hybrid architectures or distilling the planner into a smaller model, so as to preserve the benefits of hierarchical language-guided planning while approaching real-time performance.

\section{Visualization Examples and Analysis}

\begin{figure}[h]
    \centering
    \includegraphics[width=0.47\textwidth]{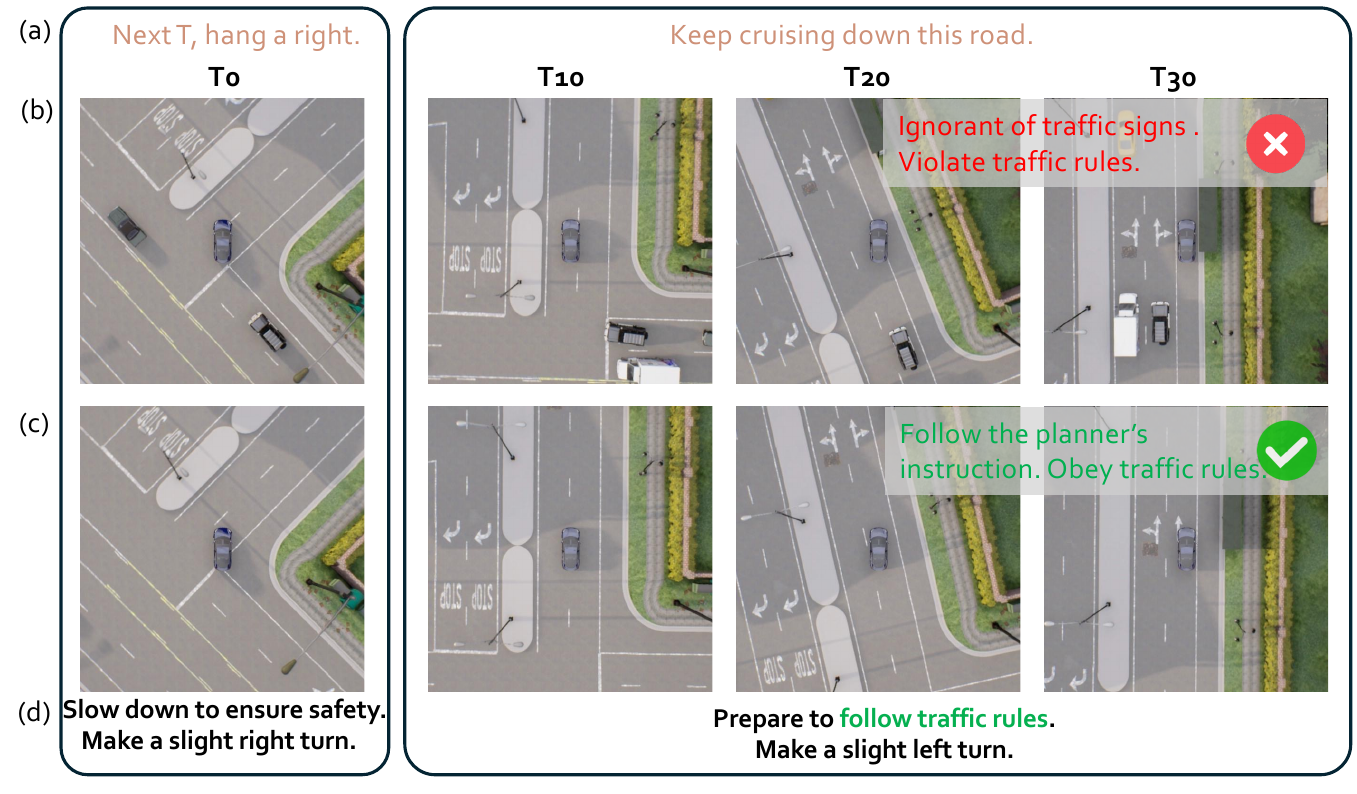}
    \caption{Comparison between AD-H and LMDrive in a multi-lane scenario.}
    \label{fig:vis_traffic}
\end{figure}

\paragraph{Traffic-compliant trajectory adjustment.}
Figure~\ref{fig:vis_traffic} compares AD-H with LMDrive in a multi-lane scenario. In the bottom row, AD-H’s planner produces traffic-compliant mid-level instructions that guide the controller to correct the trajectory and return to the appropriate lane. LMDrive drifts into the pedestrian lane (top row), revealing its limited spatial understanding and weaker generalization ability.

\paragraph{Substantial reduction of standstill behavior.}
Standstill—where the ego vehicle or robot remains stopped even under safe and navigable conditions—is a persistent challenge observed not only in autonomous driving systems~\cite{shao2023lmdrive} but also broadly in embodied Vision-language-action models~\cite{pi0-experiment-wild} that rely on learned policies. This failure mode often results from overly conservative action generation or uncertainty in policy confidence, causing disruptions in long-horizon task execution. Table~\ref{tab:standstill} reports the standstill rate on LangAuto-Tiny, measured as the proportion of episodes containing at least one standstill event. Our AD-H framework reduces the rate from \textbf{56.25\%} (LMDrive) to \textbf{12.5\%}, achieving a \textbf{43.75-percentage-point} improvement and demonstrating that the hierarchical planner–controller architecture substantially mitigates this long-standing issue.

\begin{table}[h]
\centering
\begin{tabular}{l c}
\toprule
\textbf{Method} & \textbf{Standstill Rate} \\
\midrule
LMDrive~\cite{shao2023lmdrive} & 56.3\% \\
AD-H (ours)                    & \textbf{12.5\%} \\
\bottomrule
\end{tabular}
\vspace{0.5mm}
\caption{Standstill rate on the LangAuto-Tiny benchmark. AD-H significantly reduces standstill frequency.}
\label{tab:standstill}
\end{table}

\begin{figure}[h]
    \centering
    \includegraphics[width=0.42\textwidth]{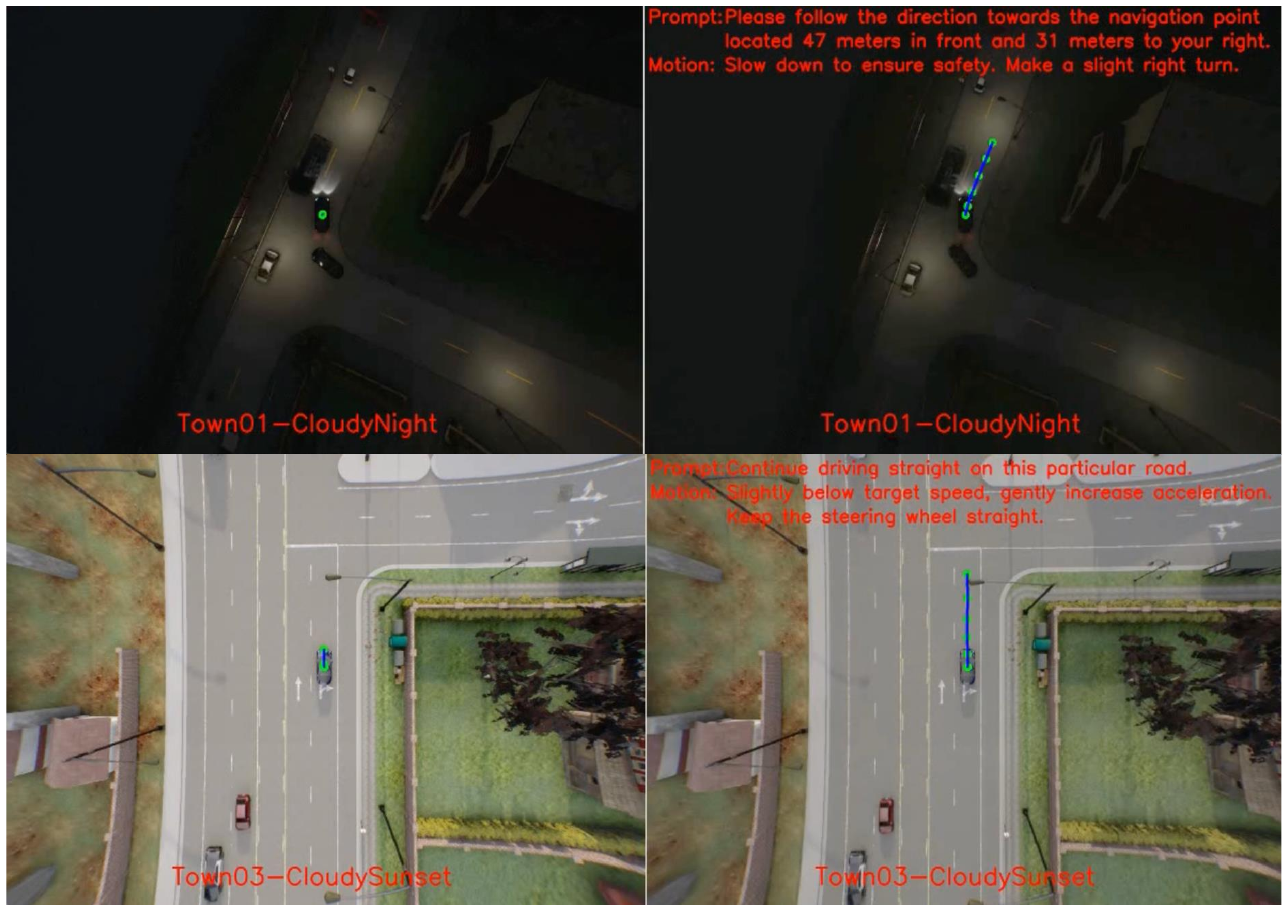}
    \caption{Examples where AD-H avoids standstill.}
    \label{fig:standstill_vis}
\end{figure}

\paragraph{Visualization of standstill-related behavior.}
Figure~\ref{fig:standstill_vis} shows scenarios where AD-H maintains safe, continuous motion, whereas LMDrive tends to stall. Although LMDrive’s planner predicts future waypoints indicating forward movement, the controller often remains stationary, reflecting an over-conservative mapping from language guidance to control. In contrast, AD-H generates consistent mid-level commands and executes them smoothly, avoiding standstill and preserving driving continuity.

\section{Conclusion and Future Work}

This work demonstrates that a hierarchical paradigm enables language-guided autonomous driving systems to better leverage the strengths of MLLM-based planners and lightweight controllers, achieving stronger generalization than end-to-end approaches. The hierarchical design yields more consistent and interpretable behaviors across varied scenarios.
However, real-world driving still contains rare and safety-critical corner cases that remain underexplored. Future work should investigate how planners interpret uncertainty, how controllers react to unexpected changes, and how to maintain stable performance under distribution shifts. Enhancing safety—through risk-aware planning, robust control, fallback behaviors, and formal safety checks—will be essential for advancing the real-world deployment of hierarchical, language-guided driving systems.

\section{Conflict of Interest}
The authors declare that they have no conflict of interest regarding the publication of this work.

\clearpage

\bibliographystyle{IEEEtran}
\bibliography{IEEEabrv,sample}

\end{document}